\documentclass[12pt]{article}
\usepackage[dvips]{color,graphicx}
\usepackage{amsmath,amssymb}
\usepackage{epsf}

\newtheorem{theorem}{Theorem}
\newtheorem{lemma}[theorem]{Lemma}
\newtheorem{corollary}[theorem]{Corollary}
\newtheorem{example}[theorem]{Example}

\title{Asymptotic Accuracy of Bayes Estimation\\
for Latent Variables with Redundancy}

\author{Keisuke Yamazaki\\
       k-yam@math.dis.titech.ac.jp \\
       Department of Computational Intelligence and Systems Science,\\
       Tokyo Institute of Technology\\
       G5-19 4259 Nagatsuta Midori-ku Yokohama, Japan
	}
\date{}
\begin{document}
\sloppy
\maketitle

\begin{abstract}
Hierarchical parametric models consisting of observable and latent variables
are widely used for unsupervised learning tasks.
For example, a mixture model is a representative hierarchical model for clustering.
From the statistical point of view,
the models can be regular or singular due to the distribution of data.
In the regular case, the models have the identifiability;
there is one-to-one relation
between a probability density function for the model expression and the parameter.
The Fisher information matrix is positive definite, and
the estimation accuracy of both observable and latent variables has been studied.
In the singular case,
on the other hand, the models are not identifiable and the Fisher matrix is not positive definite.
Conventional statistical analysis based on the inverse Fisher matrix is not applicable.
Recently, an algebraic geometrical analysis has been developed
and is used to elucidate the Bayes estimation of observable variables.
The present paper applies this analysis to latent-variable estimation
and determines its theoretical performance.
Our results clarify behavior of the convergence of the posterior distribution.
It is found that the posterior of the observable-variable estimation
can be different from the one in the latent-variable estimation.
Because of the difference, the Markov chain Monte Carlo method
based on the parameter and the latent variable
cannot construct the desired posterior distribution.
\end{abstract}

\section{Introduction}
\sloppy
Hierarchical parametric models are employed for unsupervised learning
in many data-mining and machine-learning applications.
Statistical analysis of the models plays an important role
for not only revealing the theoretical properties but also the practical applications.
For example, the asymptotic forms of the generalization error
and the marginal likelihood are used for model selection
in the maximum-likelihood and Bayes methods, respectively
\cite{Akaike,Schwarz,Rissanen}.

Parametric models generally fall into two cases:
regular and singular.
The present paper focuses on the models, 
the function of which are continuous and sufficiently smooth with respect to the parameter.
In regular cases, the Fisher information matrix is positive definite, and
there is a one-to-one relation between the parameter and the expression of the model 
as a probability density function.
Otherwise, the model is singular, and the parameter space includes singularities.
Due to these singularities, the Fisher information matrix is not positive definite,
and so the conventional analysis methods that rely on its inverse matrix are not applicable.
In this case, an algebraic geometrical approach can be used
to analyze the Bayes method \cite{Watanabe01a,Watanabe09:book}.

Hierarchical models have both observable and latent variables.
The latent variables represent the underlying structure of the model,
while the observable ones correspond to the given data.
For example, unobservable labels in clustering
are expressed as the latent variables in mixture models,
and the system dynamics of time-series data is a sequence of the variables
in hidden Markov models.
Hierarchical models thus have two estimation targets: observable and latent variables.
The well-known generalization error measures the performance of the prediction
of a future observable variable.
Combining the two model cases and the two estimation targets,
there are four estimation cases, which are summarized in Table \ref{tab:4est}.
\begin{table}[t]
\centering
\caption{Estimation classification according to the target variable and the model case}
\begin{tabular}{|c|c|c|}
\hline
Estimation Target \textbackslash Model Case & Regular Case & Singular Case \\
\hline
Observable Variable & Reg-OV estimation  & Sing-OV estimation \\
\hline
Latent Variable & Reg-LV estimation & Sing-LV estimation \\
\hline
\end{tabular}
\label{tab:4est}
\end{table}
We will use the abbreviations shown in the table to specify the target variable and the model case;
for example, Reg-OV estimation stands for estimation of the observable variable in the regular case.

In the present paper, we will investigate the asymptotic performance of the Sing-LV estimation. 
One of the main concerns in unsupervised learning is the estimation of unobservable parts
and in practical situations, the ranges of the latent variables are unknown,
which corresponds to the singular case.
The other estimation cases have already been studied;
the accuracy of the Reg-OV estimation has been clarified on the basis of the conventional analysis method,
and the results have been used for model selection criteria, such as AIC \cite{Akaike}.
The primary purpose for using the algebraic geometrical method is to analyze the Sing-OV estimation,
and the asymptotic generalization error of the Bayes method has been derived
for many models \cite{Aoyagi05,Aoyagi10,Rusakov,Yamazaki03a,Yamazaki03b,Yamazaki05e,Yamazaki05c,Zwiernik11}.
Recently, an error function for the latent-variable estimation was formalized
in a distribution-based manner, and its asymptotic form was determined
for the Reg-LV estimation of both the maximum likelihood and Bayes methods \cite{Yamazaki12a}.
Hereinafter, the estimation method will be assumed to be the Bayes method
unless it is explicitly stated otherwise.

In the Bayes estimation, parameter sampling from the posterior distribution
is an important process for practical applications.
The behavior of posterior distributions has been studied in the statistical literature.
The convergence rate of the posterior distribution has been analyzed (e.g., \cite{Ghosal+2000, LeCam1973, Ibragimov+1981}).
Specifically, the rate based on the Wasserstein metrics is elucidated
in finite and infinite mixture models \cite{Nguyen2013}.
To avoid singularities, conditions for the identifiability guaranteeing the positive Fisher matrix
are necessary.
Allman et al. use algebraic techniques to clarify the identifiability
in some hierarchical models \cite{Allman+2009}.
In the regular case, the posterior distribution has the asymptotic normality,
which means that it converges to a Gaussian distribution.
Because the variance of the distribution goes to zero when the number of data is sufficiently large,
the limit distribution is the delta distribution.
Then, the sample sequence from the posterior distribution converges to a point.
On the other hand, in the singular case,
the posterior distribution does not have the asymptotic normality
and the sequence converges to some area of the parameter space \cite{Watanabe01a}.
Studies on the Sing-OV estimation such as  \cite{Yamazaki13a} have shown that
the convergence area of the limit distribution depends on a prior distribution.
The behavior of the posterior distribution has not been clarified in the Sing-LV estimation.
The analysis of the present paper enables us to elucidate the relation between the prior and the limit posterior distributions.

The main contributions of the present paper are summarized as follows:
\begin{enumerate}
\item The algebraic geometrical method for the Sing-OV estimation
is applicable to the analysis of the Sing-LV estimation.
\item The asymptotic form of the error function is obtained, and 
its dominant order is larger than that of the Reg-LV estimation.
\item There is a case, where the limit posterior distribution in the Sing-LV estimation
is different from that in the Sing-OV estimation.
\end{enumerate}
The third result is important for practical applications:
in some priors, parameter-sampling methods based on latent variables, such as Gibbs sampling
in the Markov chain Monte Carlo (MCMC) method, 
cannot construct the proper posterior distribution
because the sample sequence of the MCMC method follows the posterior of the Sing-LV estimation,
which has a different convergence area from the desired one in the Sing-OV estimation.

The rest of this paper is organized as follows.
The next section formalizes the hierarchical model and the singular case,
and introduces the performance of the Reg-OV and the Sing-OV estimations.
Section \ref{sec:FEandPC} explains the asymptotic analysis of the free energy function
and the convergence of the posterior distribution
based on the results of the Sing-OV estimation.
In Section \ref{sec:definitions}, the latent-variable estimation
and its evaluation function are formulated in a distribution-based manner.
Section \ref{sec:main} shows the main results:
the asymptotic error function of general hierarchical models,
and the detailed error properties in mixture models.
In Section \ref{sec:Disc}, we discuss the limit distribution of the posterior in the Sing-LV estimation
and differences from the Sing-OV estimation.
Finally, Section \ref{sec:Conc} presents conclusions.
\section{The Singular Case and Accuracy of the Observable-Variable Estimation}
\label{sec:AGmethod}
In this section, we introduce the singular case
and formalize the Bayes method for the observable-variable estimation.
This section is a brief summary of the results on the Reg-OV and the Sing-OV estimations.
\subsection{Hierarchical Models and Singularities}
Let a learning model be defined by
\begin{align*}
p(x|w) =& \sum_{y=1}^K p(x,y|w) = \sum_{y=1}^K p(y|w)p(x|y,w),
\end{align*}
where $x\in R^M$ is an observable variable,
$y\in\{1,\ldots,K\}$ is a latent one,
and $w\in W \subset R^d$ is a parameter.
For the discrete $x$ such that $x\in\{1,2,\dots,M\}$,
all results hold by replacing $\int dx$ with $\sum_{x=1}^M$.
\begin{example}
\label{ex:def_mixture}
A mixture of distributions is described by
\begin{align}
p(x|w) =& \sum_{k=1}^K a_k f(x|b_k),\label{eq:mixturelearner}
\end{align}
where $f$ is the density function associated with a mixture component,
which is identifiable for any $b_k \in W_b \subset R^{d_c}$.
The mixing ratios have constraints $a_k\ge 0$ and $\sum_{k=1}^K a_k =1$.
We regard $a_1$ as a function of the parameters $a_1=1-\sum_{k=2}^K a_k$.
The parameter $w$ consists of $\{a_2,\dots,a_K\}$ and $\{b_1,\dots,b_K\}$,
where $w \in \{[0,1]^{K-1},W_b^{Kd_c}\}$.
The latent variable $y$ is the component label.
\end{example}

Assume that the number of data is $n$ and
the observable data $X^n=\{x_1,\cdots,x_n\}$ are
independent and identically distributed from
the true model, which is expressed as
\begin{align*}
q(x) =& \sum_{y=1}^{K^*} q(y)q(x|y).
\end{align*}
Note that the value range of the latent variable $y$ described as $[1,\dots,K^*]$
is generally unknown and can be different from the one in the learning model.
In the example of the mixture model,
the true model is expressed as
\begin{align}
q(x) =& \sum_{k=1}^{K^*} a^*_k f(x|b^*_k).\label{eq:mixturetrue}
\end{align}

We also assume that the true model satisfies the minimality condition:
\begin{align*}
k \ne j \in \{1,\dots,K^*\} \Rightarrow q(x|y=k)\ne q(x|y=j).
\end{align*}
For example, consider a three-component model such that $q(x|y=1)\neq q(x|y=2)=q(x|y=3)$.
This model does not satisfy the minimality condition.
Defining a new label, we obtain the following two-component expression,
which satisfies the condition;
\begin{align*}
q(x) =& q(y=1)q(x|y=1) + \{ q(y=2)+q(y=3)\}q(x|y=2)\\
=& q(y=1)q(x|y=1) + q(y=\bar{2})q(x|y=\bar{2}),
\end{align*}
where $y\in\{1,\bar{2}\}$ and $\bar{2}=\{2,3\}$.

\begin{figure}[t]
\begin{tabular}{cc}
\begin{minipage}{0.5\hsize}
\includegraphics[angle=-90,width=\columnwidth]{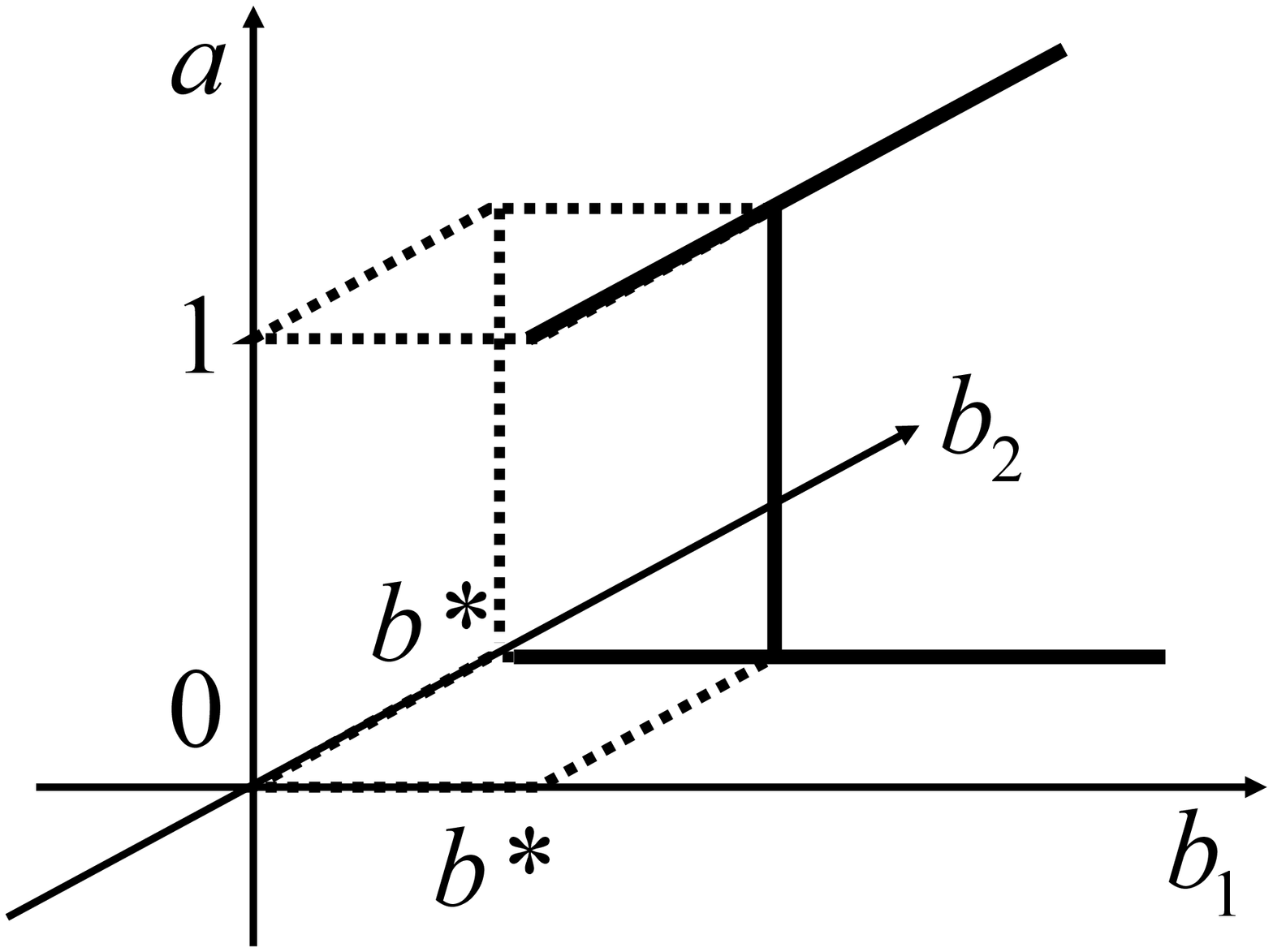}
\end{minipage}
&
\begin{minipage}{0.5\hsize}
\includegraphics[angle=-90,width=\columnwidth]{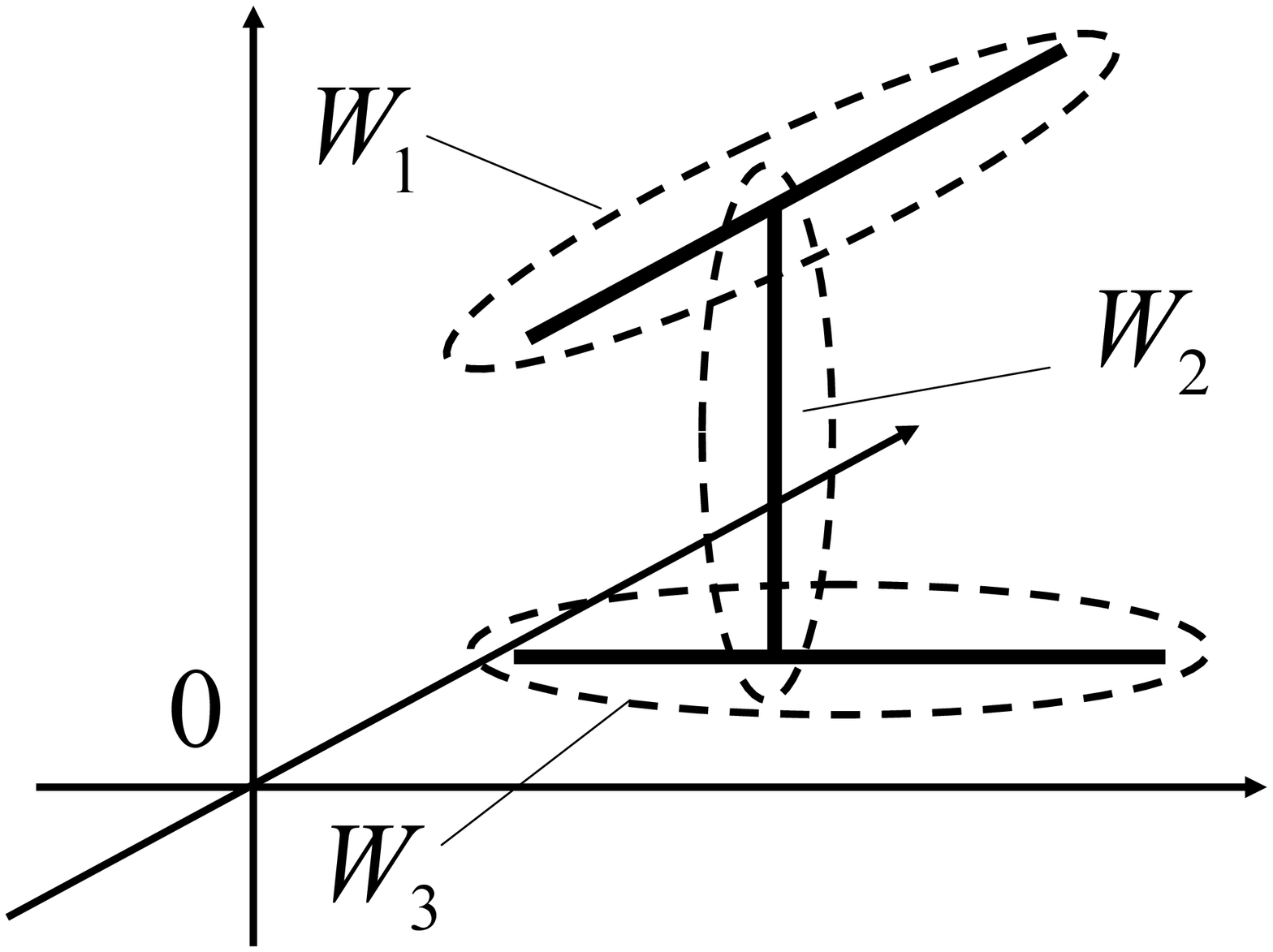}
\end{minipage}
\end{tabular}
\caption{The true parameter set $W_X^t$ (the left panel),
and the parameter areas $W_1$, $W_2$, and $W_3$ (the right panel)}
\label{fig:Areas}
\end{figure}

The present paper focuses on the case in which the true model is in the class of the learning model.
More formally, there is a set of parameters expressing the true model such that
\begin{align*}
W_X^t =& \{w^*;p(x|w^*)=q(x)\} \neq \emptyset,
\end{align*}
which is referred to as the true parameter set for $x$.
This means that the latent variable range satisfies $K=K^*$ or $K>K^*$.
The former relation corresponds to the regular case and the latter one to the singular case.
The true parameter set $W_X^t$ includes $K!$ isolated points in the regular case
due to the symmetry of the parameter space.
On the other hand, it consists of an analytic set in the singular case.
We explain this structure using the following model settings.
\begin{example}
\label{ex:2comp_gaussian}
Assume that $K=2$ and $K^*=1$ in the mixture model.
For illustrative purposes,
let the learning and the true models be defined by
\begin{align*}
p(x|w) =& af(x|b_1) +  (1-a)f(x|b_2),\\
q(x) =& f(x|b^*),
\end{align*}
respectively,
where $x\in R^1$ and $w=\{a,b_1,b_2\}$
such that $a\in [0,1]$ and $b_1,b_2\in W_b \subset R^1$.
%
We can confirm that the true parameter set consists of the following analytic set:
\begin{align*}
W_X^t =& W^t_1 \cup W^t_2 \cup W^t_3,\\
W^t_1 =& \{a=1,b_1=b^*\},\\
W^t_2 =& \{b_1=b_2=b^*\},\\
W^t_3 =& \{a=0,b_2=b^*\}.
\end{align*}
As shown in Figure \ref{fig:Areas},
let $W_1$, $W_2$, and $W_3$ be the neighborhood of $W^t_1$, $W^t_2$, and $W^t_3$, respectively.
The Fisher information matrix is not positive definite in $W^t_X$.
Moreover, the intersections of $W^t_1$, $W^t_2$ and $W^t_3$ are singularities.
\end{example}
When $K=K^*$, $W^t_X$ is a set of points, which corresponds to the regular case;
\begin{example}
\label{ex:mixture_regular}
If both the learning and the true models have two components,
\begin{align*}
p(x|w) =& af(x|b_1) +(1-a)f(x|b_2),\\
q(x) =& a^*f(x|b^*_1) +(1-a^*)f(x|b^*_2)
\end{align*}
for $a^*\neq 0,1$ and $b^*_1\neq b^*_2$,
the estimation will be in the regular case.
Due to $K!=2!=2$, the set consists of two isolated points;
\begin{align*}
W_X^t =& \{(a=a^*,b_1=b^*_1,b_2=b^*_2),(a=1-a^*,b_1=b^*_2,b_2=b^*_1)\},
\end{align*}
where the Fisher information matrix is positive definite.
\end{example}
\subsection{The Observable-Variable Estimation and its Performance}
In Bayesian statistics, estimation of the observable variables is defined by
\begin{align*}
p(x|X^n) =& \int p(x|w)p(w|X^n)dw ,\\
p(w|X^n) =& \frac{\prod_{i=1}^np(x_i|w)\varphi(w;\eta)}{Z(X^n)},
\end{align*}
where $\varphi(w;\eta)$ is a prior distribution with the hyperparameter $\eta$,
$p(w|X^n)$ is the posterior distribution of the parameter,
and its normalizing factor is given by
\begin{align*}
Z(X^n) =& \int \prod_{i=1}^np(x_i|w)\varphi(w;\eta)dw.
\end{align*}
This formulation is available for both the Reg-OV and Sing-OV estimations.
In the mixture model, the Dirichlet distribution is often used
for the prior distribution of the mixing ratio;
\begin{align}
\varphi(w;\eta) =& \varphi(a;\eta_1)\varphi(b;\eta_2),\label{eq:mixtureprior}\\
\varphi(a;\eta_1) =& \frac{\Gamma(K\eta_1)}{\Gamma(\eta_1)^K}\prod_{i=k}^K a_k^{\eta_1-1},\label{eq:mixtureprior_a}
\end{align}
where $a=\{a_1,\dots,a_K\}$, $b=\{b_1,\dots,b_K\}$, $\eta=\{\eta_1,\eta_2\}\in R^2_{>0}$, and
$\Gamma$ is the gamma function.
Since $a_k$ has the same exponential part for all $k$, 
$\varphi(a;\eta_1)$ is referred to as a symmetric Dirichlet distribution.

The estimation accuracy is measured by the average Kullback-Leibler divergence:
\begin{align*}
G(n) =& E_X\bigg[ \int q(x) \ln \frac{q(x)}{p(x|X^n)}dx \bigg],
\end{align*}
where the expectation is
\begin{align*}
E_X[f(X^n)] =& \int f(X^n) q(X^n)dX^n.
\end{align*}

Let us define the free energy as
\begin{align*}
F(X^n) =& -\ln Z(X^n),
\end{align*}
which plays an important role in Bayes statistics as a criterion for selecting the optimal model.
In the Reg-OV estimation,
the Bayesian information criterion (BIC; \cite{Schwarz})
 and the minimum-description-length principle (MDL; \cite{Rissanen}) are both based on the asymptotic form of $F(X^n)$.
Theoretical studies often analyze the average free energy given by
\begin{align*}
F_X(n) =& -nS_X + E_X[F(X^n)],
\end{align*}
where the entropy function is defined by
\begin{align*}
S_X =& -\int q(x)\ln q(x)dx.
\end{align*}
The model that minimizes $F(X^n)$ is then selected as optimal
from among the candidate models.
The energy function $F_X(n)$ allows us to investigate the average behavior of the selection.
Note that the entropy term does not affect the selection result
because it is independent of the candidate models.
According to the definitions,
the average free energy and the generalization error have the relation
\begin{align}
G(n) =& E_{X^n}\bigg[ \int q(x_{n+1})\ln\frac{q(x_{n+1})}{p(x_{n+1}|X^n)}dx_{n+1}\bigg]\nonumber\\
=& E_{X^n,x_{n+1}}\bigg[ \ln \frac{q(x_{n+1})}{p(x_{n+1}|X^n)}\bigg]\nonumber\\
=& E_{X^n,x_{n+1}}\bigg[ \ln \frac{\prod_{i=1}^{n+1} q(x_i)}{\int \prod_{i=1}^{n+1} p(x_i|w)\varphi(w;\eta)dw}\bigg]\nonumber\\
&-E_{X^n}\bigg[ \ln \frac{\prod_{i=1}^n q(x_i)}{\int \prod_{i=1}^n p(x_i|w)\varphi(w;\eta)dw}\bigg]\nonumber\\
=& F_X(n+1) - F_X(n), \label{eq:GFF}
\end{align}
which implies that the asymptotic form of $F(n)$ also relates to that of $G(n)$.
The rest of the paper discusses the case $W_X^t\neq \emptyset$,
although it is also important to consider the case $W_X^t=\emptyset$, where the learning model
cannot attain the true model.

The algebraic geometrical analysis \cite{Watanabe01a,Watanabe09:book}
is applicable to both the regular and singular cases for deriving the asymptotic form of $F_X(n)$.
Its result shows that the form is expressed as
\begin{align*}
F_X(n) =&  \lambda_X\ln n -(m_X-1)\ln\ln n +O(1),
\end{align*}
where the coefficients $\lambda_X$ and $m_X$ are
positive rational and natural, respectively.
The reason why the free energy has this form will be explained in the next section.
According to the relation Eq.\ref{eq:GFF}, the asymptotic form of the generalization error is given by
\begin{align}
G(n) =& \frac{\lambda_X}{n} - \frac{m_X-1}{n \ln n} + o\bigg(\frac{1}{n \ln n}\bigg).\label{eq:asympGn}
\end{align}
Since the learning model can attain the true model,
we can confirm that the generalization error converges to zero for $n\rightarrow\infty$.
The coefficients are $\lambda_X=d/2$ and $m_X=1$ in the regular case.
It is proved that $\lambda_X<d/2$ in the singular case (Section 7 in \cite{Watanabe09:book}).
\section{Asymptotic Analysis of the Free Energy and Posterior Convergence}
\label{sec:FEandPC}
This section introduces the asymptotic analysis of $F_X(n)$ based on algebraic geometry
and explains how the prior distribution affects convergence of the posterior distribution.
The topics in this section have already been elucidated in the studies on the Sing-OV estimation
(e.g., \cite{Watanabe09:book}).
\subsection{Relation between the Free Energy and the Zeta Function}
Let us define another Kullback-Leibler divergence,
\begin{align*}
H_X(w) =& \int q(x) \ln \frac{q(x)}{p(x|w)}dx,
\end{align*}
which is assumed to be analytic (Fundamental Condition I in \cite{Watanabe09:book}).
We consider the prior distribution $\varphi(w;\eta)=\psi_1(w;\eta)\psi_2(w;\eta)$,
where $\psi_1(w;\eta)$ is a positive function of class $C^{\infty}$
and $\psi_2(w;\eta)$ is a nonnegative analytic function (Fundamental Condition II in \cite{Watanabe09:book}).
Let the zeta function of a parametric model be given by
\begin{align*}
\zeta_X(z) =& \int H_X(w)^z \varphi(w;\eta)dw,
\end{align*}
where $z$ is a complex variable.
From algebraic analysis, we know that its poles are real, negative, and rational \cite{Atiyah}.
Let the largest pole and its order be $z=-\lambda_X$ and $m_X$, respectively.
The zeta function includes the term
\begin{align*}
\zeta_X(z) =& \frac{f_c(z)}{(z+\lambda_X)^{m_X}}+\dots,
\end{align*}
where $f_c(z)$ is a holomorphic function.
We define the state density function of $t>0$ as
\begin{align*}
v(t) =& \int \delta(t-H_X(w))\varphi(w;\eta)dw.
\end{align*}
The zeta function is its Mellin transform:
\begin{align*}
\zeta_X(z) =& \mathcal{M}[v(t)]=\int_0^\infty v(t)t^z dt.
\end{align*}
Moreover, it is known that the inverse Laplace transform of $v(t)$
has the same asymptotic form as $F_X(n)$;
\begin{align*}
\mathcal{L}^{-1}[v(t)] =& \int v(t)e^{nt}dt\\
=& \int e^{nH_X(w)}\varphi(w;\eta)dw=F_X(n).
\end{align*}
Then, there is the following relation,
\begin{align*}
F_X(n) \overset{\mathcal{L}}{\Longleftrightarrow} v(t) 
\overset{\mathcal{M}}{\Longleftrightarrow} \zeta_X(z).
\end{align*}
Based on the Laplace and the Mellin transforms,
the asymptotic forms of all functions are available
if one of them is given.
Following the transforms from $\zeta_X(z)$ to $F_X(n)$ through $v(t)$,
we obtain the asymptotic form
\begin{align*}
F_X(n) =& \lambda_X\ln n -(m_X-1)\ln\ln n +O(1).
\end{align*}

Let us define the effective area of the parameter space,
which plays an important role in the convergence analysis of the posterior distribution.
According to the results on the Sing-OV estimation,
it has been found that the largest pole exists in a restricted parameter space.
In Example \ref{ex:2comp_gaussian}, the parameter space is divided into
$W_1$, $W_2$, $W_3$ and the rest of the support of $\varphi(w;\eta)$.
The first three sets are neighborhoods of the analytic sets $W^t_1$, $W^t_2$ and $W^t_3$
constructing $W^t_X$, respectively.
Assume that a pole $z=-\lambda_e$ of the zeta function
\begin{align*}
\zeta_e(z)=& \int_{W_e} H_X(w)^z\varphi(w;\eta)dw
\end{align*}
is equal to the largest pole $z=-\lambda_X$, where $W_e=W_1\cap W_2$.
In the present paper, we refer to $W_e$ as \emph{the effective area}.
Let the effective area be denoted by the minimum set $W_1\cap W_2$.
In other words, we do not call $W_1$ the effect area even though $W_1$ includes $W_e$.
If the largest pole of $\int_{W_1\setminus W_1\cap W_2} H_X(w)^z\varphi(w;\eta)dw$
is also equal to $z=-\lambda_X$,
the effective area is $W_1$ since $W_1\cap W_2$ can not cover the area.
\subsection{Phase Transition}
A switch in the underlying function of the free energy is generally referred to as a phase transition.
When the prior of the mixing ratio parameters is the Dirichlet distribution.
the phase transition is observed in $F_X(n)$.
Combining the results of \cite{Yamazaki10a} and \cite{Yamazaki13a},
we obtain the following lemma;
\begin{lemma}
\label{lem:binomial_D}
Suppose that $K=2$, $K^*=1$ in the mixture model, where the true and the learning models are given by
\begin{align*}
q(x) =& f(x|b^*),\\
p(x|w) =& af(x|b_1)+(1-a)f(x|b_2),
\end{align*}
respectively.
Let the component be expressed as
\begin{align*}
f(x=m|b_k) =& \binom{M}{m} b_k^m(1-b_k)^{M-m},
\end{align*}
where $x\in\{1,\dots,M\}$,
$M$ is an integer such that $K<M$,
and $(M\; m)^\top$ is the binomial coefficient.
We consider the case $0<b^*<1$.
Let the prior distribution for the mixing ratio be the symmetric Dirichlet distribution,
and the one for $b_k$ be analytic and positive.
Then the largest pole of the zeta function $\zeta_X(z)$ is
\begin{align*}
\lambda_X =& 
\begin{cases}
\frac{1+\eta_1}{2} & \eta_1 \le 1/2,\\
\frac{3}{4} & \eta_1 > 1/2,
\end{cases}\\
m_X =&
\begin{cases}
2 & \eta_1 = 1/2,\\
1 & \text{otherwise}.
\end{cases}
\end{align*}
Moreover, the effective area $W_e$ is given by
\begin{align*}
W_e =& 
\begin{cases}
W_1 \cup W_3 & \eta_1<1/2,\\
(W_1\cap W_2)\cup(W_3\cap W_2) & \eta_1=1/2,\\
W_2 & \eta_1>1/2.
\end{cases}
\end{align*}
\end{lemma}
The proof is in the Appendix C.
Lemma \ref{lem:binomial_D} indicates that the free energy has the phase transition at $\eta_1=1/2$.
\subsection{Convergence Area of the Posterior Distribution}
The asymptotic form of the free energy determines the limit structure of the posterior distribution.
In this subsection, we will show that the convergence area is the effective parameter area.

The free energy $F(X^n)$ has an asymptotic form similar to
the average energy $F_X(n)$ (Main Formula II in \cite{Watanabe09:book}),
\begin{align}
F(X^n) =& nS(X^n)+\lambda_X \ln n -(m_X-1)\ln\ln n + O_p(1), \label{eq:asymp_FX}
\end{align}
where $S(X^n)=\frac{1}{n}\sum_{i=1}^n\ln q(x_i)$.
According to $Z(X^n) = \exp(-F(X^n))$, the posterior distribution has the expression,
\begin{align*}
p(w|X^n) =& \frac{\prod_{i=1}^n p(x_i|w)\varphi(w;\eta)}{\exp\{-nS(X^n)-\lambda_X \ln n +o_p(\ln n)\}}.
\end{align*}
Let us divide the neighborhood of $W_X^t$ into $W_e \cup W_o$,
where $W_e$ is the effective area.
Then, there is a pole $z=-\mu_X$ such that $\mu_X>\lambda_X$ in the other area $W_o$,
and the posterior value of $W_o$ is described by
\begin{align*}
p(W_o|X^n) =& \int_{W_o} p(w|X^n) dw\\
=& \frac{\int_{W_o}\prod_{i=1}^n p(x_i|w)\varphi(w;\eta)dw}{\exp\{-nS(X^n)-\lambda_X \ln n +o_p(\ln n)\}}\\
=& \frac{\exp\{-nS(X^n)-\mu_X \ln n +o_p(\ln n)\}}{\exp\{-nS(X^n)-\lambda_X \ln n +o_p(\ln n)\}}\\
=& n^{-\mu_X+\lambda_X} +o_p(n^{-\mu_X+\lambda_X}).
\end{align*}
The posterior asymptotically has zero value in $W_o$,
which means that it converges to the effective area.

According to Lemma \ref{lem:binomial_D},
the effective area depends on the hyperparameter.
Therefore, the convergence area changes at the phase transition point $\eta_1=1/2$.
It also shows how the learning model realizes the true one.
In $W_1\cup W_3$, the true model is expressed by one-component model,
which means that the redundant component is eliminated.
On the other hand, all components of the learning model are used in $W_2$.

The phase transition is observed in general mixture models;
\begin{theorem}
\label{th:PT_mixture}
Let a learning model and the true one be expressed
as Eqs \ref{eq:mixturelearner} and \ref{eq:mixturetrue}, respectively.
When the prior of the mixing ratio is the Dirichlet distribution of Eq.~\ref{eq:mixtureprior_a},
the average free energy $F_X(n)$ has at least two phases:
the phase that eliminates all redundant components when $\eta_1$ is small,
and the one that uses them when $\eta_1$ is sufficiently large.
\end{theorem}
The proof is in the Appendix C.
\section{Formal Definition of the Latent-Variable Estimation and its Accuracy}
\label{sec:definitions}
This section formulates the Bayes latent-variable estimation
and an error function that measures its accuracy.

We first consider a detailed definition of a latent variable.
Let $Y^n=\{y_1,\ldots,y_n\}$ be unobservable data,
which correspond to the latent parts of the observable $X^n$.
Then, the complete form of the data is $(x_i,y_i)$, and
$(X^n,Y^n)$ and $X^n$ are referred to as complete and incomplete data, respectively.
The true model generates the complete data $(X^n,Y^n)$,
where the range of the latent variables is $y_i\in\{1,\dots,K^*\}$.
The learning model, on the other hand, has the range $y_i\in\{1,\dots,K\}$.
For a unified description, we define that the true model has probabilities
$q(y)=0$ and $q(x,y)=0$ for $y>K^*$.

We define the true parameter set for $(x,y)$ as
\begin{align*}
W_{XY}^t = \{w^*;p(x,y|w^*)=q(x,y)\},
\end{align*}
which is a proper subset of $W_X^t$.
In Example \ref{ex:2comp_gaussian},
\begin{align*}
W_{XY}^t =& \{a=1,b_1=b^*\}=W^t_1\subset W^t_X.
\end{align*}
The subsets $W_2=\{b_1=b_2=b^*\}$ and $W_3=\{a=0,b_2=b^*\}$ in $W_X^t$ are excluded
since $W_{XY}^t$ takes account of the representation with respect to not only $x$ but also $y$.
Due to the assumption $W_X^t\neq \emptyset$, $W_{XY}^t$ is not empty.
The set $W_{XY}^t$ again consists of an analytic set in the singular case,
and it is a unique point in the regular case.

While latent-variable estimation falls into various types according to the target of the estimation,
the present paper focuses on the Type-I estimation of \cite{Yamazaki12a}:
the joint probability of $(y_1,\ldots,y_n)$ is the target and is
written as $p(Y^n|X^n)$.
The Bayes estimation has two equivalent definitions:
\begin{align}
p(Y^n|X^n) =& \int \prod_{i=1}^n \frac{p(x_i,y_i|w)}{p(x_i|w)}p(w|X^n)dw \label{eq:learner_posterior}\\
=& \frac{Z(X^n,Y^n)}{Z(X^n)},\label{eq:learner}
\end{align}
where the marginal likelihood for the complete data is given by
\begin{align*}
Z(X^n,Y^n) =& \int \prod_{i=1}^n p(x_i,y_i|w)\varphi(w;\eta)dw.
\end{align*}
It is easily confirmed that $Z(X^n)=\sum_{Y^n} Z(X^n,Y^n)$.

The true probability of $Y^n$ is uniquely given by
\begin{align}
q(Y^n|X^n) =& \frac{q(X^n,Y^n)}{q(X^n)} = \prod_{i=1}^n \frac{q(x_i,y_i)}{q(x_i)}.\label{eq:true}
\end{align}
The accuracy of the estimation is measured by the difference between
$q(Y^n|X^n)$ and $p(Y^n|X^n)$.
Thus, we define the error function as the average Kullback-Leibler divergence,
\begin{align}
D(n) =& \frac{1}{n}E_{XY}\bigg[ \ln\frac{q(Y^n|X^n)}{p(Y^n|X^n)} \bigg],\label{eq:def_Dn}
\end{align}
where the expectation is defined as
\begin{align*}
E_{XY}[f(X^n,Y^n)] =& \int \sum_{y_1=1}^{K}\cdots\sum_{y_n=1}^{K} f(X^n,Y^n) q(X^n,Y^n)dX^n.
\end{align*}
\section{Asymptotic Analysis of the Error Function}
\label{sec:main}
In this section, we show that the algebraic geometrical analysis is applicable
to the Sing-LV estimation, and present the asymptotic form of the error function $D(n)$.

\subsection{Conditions for the Analysis}
Before showing the asymptotic form of the error function,
we state necessary conditions.

Let us define the zeta function on the complete data $(x,y)$ as
\begin{align*}
\zeta_{XY}(z) =& \int H_{XY}(w)^z\varphi(w;\eta)dw,
\end{align*}
where the Kullback-Leibler divergence $H_{XY}(w)$ is given by
\begin{align*}
H_{XY}(w) =& \sum_{y=1}^{K}\int q(x,y)\ln \frac{q(x,y)}{p(x,y|w)}dx.
\end{align*}
Let the largest pole of $\zeta_{XY}(z)$ be $z=-\lambda_{XY}$, and let its order be $m_{XY}$.

We consider the following conditions:
\begin{description}
\item[(A1)] The divergence functions $H_{XY}(w)$ and $H_X(w)$ are analytic.
\item[(A2)] The prior distribution has the compact support,
which includes $W_X^t$, and has the expression $\varphi(w;\eta)=\psi_1(w;\eta)\psi_2(w;\eta)$,
where $\psi_1(w;\eta)>0$ is a function of class $C^{\infty}$ and $\psi_2(w;\eta)\ge 0$ is analytic
on the support of $\varphi(w;\eta)$.
\end{description}
They correspond to the Fundamental Conditions I and II in \cite{Watanabe09:book}, respectively.
It is known that models with discrete $x$ such as the binomial mixture satisfy (A1) \cite{Yamazaki10a}.
On the other hand, if $x$ is continuous, there are some models, of which $H_X(w)$ is not analytic;
\begin{example}[Example 7.3 in \cite{Watanabe09:book}]
\label{ex:non_analytic}
In the Gaussian mixture, $H_X(w)$ is not analytic, which means that the mixture model does not satisfies (A1).
Let us consider a simple case; $K=2$ and $K^*=1$,
where the true model and a learning model are given by
\begin{align*}
q(x) =& f(x|0),\\
p(x|a) =& a f(x|2) +(1-a)f(x|0),
\end{align*}
respectively, where $x \in R^1$,
\begin{align*}
f(x|b) =& \frac{1}{\sqrt{2\pi}}\exp\bigg\{-\frac{(x-b)^2}{2}\bigg\},
\end{align*}
and $b\in W^1=R^1$.
Then,
\begin{align*}
H_X(a) =& \int q(x) \ln\frac{q(x)}{p(x|a)}dx \\
=& - \int q(x) \ln\{1+a(\exp(2x-2)-1)\}dx \\
=& \int \sum_{j=1}^\infty \frac{a^j}{j}(1-\exp(2x-2))^j q(x)dx,
\end{align*}
where the last expression is a formal expansion.
Since its convergence radius is zero at $a=0$, $H_X(a)$ is not analytic.
Based on the similar way, we can find that $H_X(w)$ is not analytic in a general Gaussian mixture .
\end{example}
The following example shows a prior distribution for the mixture model satisfying (A2).
\begin{example}
\label{ex:prior_mixture}
The symmetric Dirichlet distribution satisfies the condition (A2)
because Eq.~\ref{eq:mixtureprior_a} is obviously analytic and non negative in its support.
Choosing an analytic distribution for $\varphi(b;\eta_2)$,
we obtain the prior $\varphi(w;\eta)$ satisfying the condition (A2).
\end{example}
\subsection{Asymptotic Form of the Error Function}
Now, we show the main theorem on the asymptotic form of the error function:
\begin{theorem}
\label{th:main}
Let the true distribution of the latent variables and the estimated distribution
be defined by Eqs. \ref{eq:true} and \ref{eq:learner}, respectively.
By assuming the conditions (A1) and (A2),
the asymptotic form of $D(n)$ is expressed as
\begin{align*}
D(n) =& (\lambda_{XY}-\lambda_X)\frac{\ln n}{n} - (m_{XY}-m_X)\frac{\ln\ln n}{n} + o\bigg(\frac{\ln\ln n}{n}\bigg).
\end{align*}
\end{theorem}
The proof is in Appendix A.
The theorem indicates that the algebraic geometrical method plays an essential role
for the analysis of the Sing-LV estimation
because the coefficients consist of the information of the zeta functions
such as $\lambda_{XY}$, $\lambda_X$, $m_{XY}$ and $m_X$.
The order $\ln n/n$ has not ever appeared in the Reg-LV estimation.
In the Reg-LV estimation such that $K=K^*$,
the asymptotic error function has the following form \cite{Yamazaki12a};
\begin{align*}
D(n) =& \frac{1}{n}\mathrm{Tr}[I_{XY}(w^*)I_X(w^*)^{-1}] + o\bigg(\frac{1}{n}\bigg),\\
\{I_{XY}(w)\}_{ij} =& \sum_{y=1}^K \int
\frac{\partial \ln p(x,y|w)}{\partial w_i}\frac{\partial \ln p(x,y|w)}{\partial w_j} p(x,y|w)dx,\\
\{I_X(w)\}_{ij} =& \int
\frac{\partial \ln p(x|w)}{\partial w_i}\frac{\partial \ln p(x|w)}{\partial w_j} p(x|w) dx,\\
\end{align*}
where $w^*$ is the unique point consisting of $W_{XY}^t$.
The dominant order is $1/n$, and the coefficient is determined
by the Fisher information matrices on $p(x,y|w)$ and $p(x|w)$.
Theorem \ref{th:main} implies that the largest possible order is $\ln n/ n$ in the Sing-LV estimation.
This order change is adverse for the performance
because the error converges more slowly to zero.
In singular cases,
the probability $p(Y^n|X^n)$ is constructed over the space $Y^n \in K^n$
while the true probability $q(Y^n|X^n)$ is over $Y^n\in K^{*n}$.
The size of the redundant space $K^n-K^{*n}$ grows exponentially
with the amount of training data.
For realizing $p(x,y|w^*)$, where $w^* \in W_{XY}^t$,
we must assign zero to the probabilities on the vast redundant space.
The increased order reflects the cost of assigning these values.

Let us compare the dominant order of $D(n)$ with that of the generalization error.
We find that both Reg-OV and Sing-OV estimations have the same dominant order $1/n$ as shown in Eq. \ref{eq:asympGn}
while the redundancy and the hyperparameter affect the coefficients.
Thus, changing the order is a unique phenomenon of the latent-variable estimation.
%
\subsection{Asymptotic Error in the Mixture Model}
\label{sec:case_mixture}
In Theorem \ref{th:main}, the possible dominant order was calculated as $\ln n/n$.
However, there is no guarantee that this is the actual maximum order;
the order can decrease to $1/n$ if the coefficients are zero,
where the zeta functions $\zeta_{XY}(z)$ and $\zeta_X(z)$
have their largest poles in the same position and their multiple orders are also the same.
The result of the following theorem clearly shows that the dominant order is $\ln n/n$
in the mixture models.
\begin{theorem}
\label{th:error_mixture}
Let the learning and the true models be mixtures
defined by Eqs. \ref{eq:mixturelearner} and \ref{eq:mixturetrue},
respectively.
Assume the conditions (A1) and (A2).
The Bayes estimation for the latent variables, Eq. \ref{eq:learner},
with the prior represented by Eqs. \ref{eq:mixtureprior} and \ref{eq:mixtureprior_a}
has the following bound for the asymptotic error:
\begin{align*}
D(n) \ge& \frac{(K-K^*)\eta_1}{2}\frac{\ln n}{n} + o\bigg(\frac{\ln n}{n}\bigg).
\end{align*}
\end{theorem}
The proof is in Appendix A.
Due to the definition of the Dirichlet distribution,
$\eta_1$ is positive.
Combining this with the assumption $K^*<K$,
we obtain that the coefficient of $(\ln n)/n$ is positive,
which indicates that it is the dominant order.

The Dirichlet prior distribution for the mixing ratio is qualitatively known to have
a function controlling the number of available components, the so-called automatic relevance determination (ARD);
a small hyperparameter tends to have a result with few components due to the shape of the distribution.
Theorem \ref{th:error_mixture} quantitatively shows an effect of the Dirichlet prior.
The lower bound in the theorem mathematically supports the ARD effect;
the redundancy $K-K^*$ and the hyperparameter $\eta_1$ have a linear influence on the accuracy.

Theorem \ref{th:error_mixture} holds in a wider class of the mixture models
since the error is evaluated as the lower bound.
The following corollary shows that the Gaussian mixture has the same bound
for the error even though it does not satisfy (A1) as shown in Example \ref{ex:non_analytic}.
\begin{corollary}
\label{cor:bound_gaussian}
Assume that in a mixture model, $H_{XY}(w)$ is analytic,
and the prior distribution for the mixing ratio is the symmetric Dirichlet distribution.
If there is a positive constant $C_1$ such that
\begin{align*}
H_X(w) \le& C_1\int \bigg(\frac{p(x|w)}{q(x)}-1\bigg)^2 dx,
\end{align*}
the error function has the same lower bound as Theorem \ref{th:error_mixture}.
In the Gaussian mixture, components of which are defined by
\begin{align*}
f(x|b) =& \frac{1}{\sqrt{2\pi}^M}\exp\bigg\{-\frac{||x-b||^2}{2}\bigg\},
\end{align*}
where $x\in R^M$ and $b\in W^M=R^M$,
$H_{XY}(w)$ is analytic and the inequality holds.
\end{corollary}
The proof is in Appendix A.
\section{Discussion}
\label{sec:Disc}
Theorem \ref{th:main} shows that the asymptotic error has the coefficient $\lambda_{XY}-\lambda_X$,
which is the difference of the largest poles in the zeta functions.
Based on the free energy of the complete data defined as $F(X^n,Y^n)=-\ln Z(X^n,Y^n)$,
we find that the error is determined by the different properties between $F(X^n,Y^n)$ and $F(X^n)$
since their asymptotic forms are expressed as
\begin{align*}
F(X^n,Y^n) =& nS(X^n,Y^n) + \lambda_{XY} \ln n -(m_{XY}-1)\ln \ln n +O_p(1),\\
F(X^n) =& nS(X^n) +\lambda_X \ln n -(m_X-1)\ln \ln n + O_p(1),
\end{align*}
where $S(X^n,Y^n) = -\frac{1}{n}\sum_{i=1}^n \ln q(x_i,y_i)$.

In this section, we examine the properties of $F(X^n,Y^n)$
and indicate that the difference from those of $F(X^n)$ affects the behavior of
the Sing-LV estimation and the parameter sampling from the posterior distribution.
\subsection{Effect to Eliminate Redundant Labels}
According to Eq. \ref{eq:learner},
the MCMC sampling of the $Y^n$'s following $p(Y^n|X^n)$ is essential for the Bayes estimation.
The following relation indicates that we do not need to calculate $Z(X^n)$
and that the value of $Z(X^n,Y^n)$ determines the properties of the estimation:
\begin{align}
p(X^n,Y^n)=Z(X^n,Y^n) \propto p(Y^n|X^n) = \frac{Z(X^n,Y^n)}{Z(X^n)}. \label{eq:pZZ}
\end{align}
The expression of $p(X^n,Y^n)$ can be tractable with a conjugate prior,
which marginalizes out the parameter integral \cite{Dawid1993,Heckerman1999}.

We determine where the estimated distribution $p(Y^n|X^n)$ has its peak.
Obviously, the label assignment $Y^n$ minimizing $F(X^n,Y^n)$ provides the peak
due to the definition $F(X^n,Y^n)=-\ln Z(X^n,Y^n)$ and Eq. \ref{eq:pZZ}.
Let this assignment be described as $\bar{Y}^n$;
\begin{align*}
\bar{Y}^n = \arg\max_{Y^n} p(Y^n|X^n) = \arg \min_{Y^n} F(X^n,Y^n).
\end{align*}
The following discussion shows that $\bar{Y}^n$ does not include the redundant labels.

We have to consider the symmetry of the latent variable in order to discuss the peak.
In latent-variable models, both the latent variable and the parameter are symmetric.
In Example \ref{ex:2comp_gaussian},
the component $f(x|b^*)$ of the true model can be attained by
the first component $a_1f(x|b_1)$ or the second one $(1-a_1)f(x|b_2)$ of the learning model.
Because the true label $y=1$, which the true model provides, is unobservable,
there are two proper estimation results $Y^n=\{1,\dots,1\}$ and $Y^n=\{2,\dots,2\}$
to indicate that the true model consists of one component.
This is the symmetry of the latent variable.
In the parameter space, it corresponds to the symmetric structure of $W_1$ and $W_3$ shown in Fig.\ref{fig:Areas}.
The symmetry makes it difficult to interpret the estimation results,
which is known as the label-switching problem.

For the purpose of the theoretical evaluation,
the definition of the error function $D(n)$ selects the true assignment of the latent variable.
In the above example, only $Y^n=\{1,\dots,1\}$ is accepted as the proper result.
However, there is no selection of the true assignment in the estimation process;
other symmetric assignments such as $Y^n=\{2,\dots,2\}$ will be the peak of $p(Y^n|X^n)$.
Then, the true parameter area $W^t_{XY}$ is not sufficient to describe the peak.
Taking account of the symmetry, we define another analytic set of the parameter as
\begin{align*}
W_{XY}^p =& \cup_{\sigma\in \Sigma}\{w; a_{\sigma(k)}=a^*_k, b_{\sigma(k)}=b^*_k
\;\text{for}\; 1\le k\le K^*\},
\end{align*}
$\Sigma$ is the set of injective functions from $\{1,\dots,K^*\}$ to $\{1,\dots,K\}$.
It is easy to confirm that $W^t_{XY}\subset W_{XY}^p$.
In Example \ref{ex:2comp_gaussian}, $W^t_{XY}=W^t_1\subset W^t_1\cup W^t_3=W_{XY}^p$.
Note that the redundant components are eliminated in $p(x|w^*)$, where $w^*\in W^p_{XY}$.

Let us analyze the location of the peak.
Define that
\begin{align*}
S'(X^n,Y^n) =& -\frac{1}{n} \sum_{i=1}^n \ln p(x_i,y_i|w^*),
\end{align*}
where $w^* \in W^p_{XY}$.
Switching the label based on the symmetry,
we can easily prove that $\max_{w^*,Y^n}S'(X^n,Y^n)=\max_{Y^n}S(X^n,Y^n)$.
Moreover, $-\frac{1}{n}\sum_{i=1}^n \ln p(x_i,y_i|w)$ with $w\in W_X^t \setminus W_{XY}^p$,
such as $w\in W^t_2$ in Example \ref{ex:2comp_gaussian}, cannot realize $S(X^n,Y^n)$
according to a simple calculation as shown in the next paragraph.
Because the leading term of the asymptotic $F(X^n,Y^n)$ is $nS(X^n,Y^n)$
and $nS'(X^n,\bar{Y}^n)$ realizes it,
the true assignment $\bar{Y}^n$ follows the parameter $w^* \in W^p_{XY}$.
Recalling that the redundant components are eliminated when $w\in W^p_{XY}$,
we can conclude that the redundant labels are eliminated in $\bar{Y}^n$.
This elimination occurs in any prior distribution
if its support includes $W^p_{XY}$.

Let us confirm the elimination in Example \ref{ex:2comp_gaussian}.
We consider three parameters; $w^*_1\in W^t_1=\{a=1,b_1=b^*\}$, 
$w^*_2\in W^t_2=\{b_1=b_2=b^*\}$ and $w^*_3\in W^t_3=\{a=0,b_2=b^*\}$.
The leading term of the asymptotic $F(X^n,Y^n)$ is expressed as
\begin{align*}
nS'_j(X^n,Y^n) =& -\sum_{i=1}^n \ln p(x_i,y_i|w^*_j)
\end{align*}
for $j=1,2,3$.
This is rewritten as
\begin{align}
nS'_j(X^n,Y^n) =& -\sum_{i=1}^n \delta_{y_i,1}\ln a - \sum_{i=1}^n \delta_{y_i,2}\ln (1-a)\nonumber\\
&- \sum_{i=1}^n \delta_{y_i,1}\ln f(x_i|b_1) - \sum_{i=1}^n \delta_{y_i,2}\ln f(x_i|b_2),\label{eq:S'_j}
\end{align}
where $\delta_{i,j}$ is the Kronecker delta.
The assignment $\bar{Y}^n$ depends on $w^*_j$.
For example, $\bar{Y}^n=\{1,\dots,1\}$ for $w^*_1$ and  $\bar{Y}^n=\{2,\dots,2\}$ for $w^*_3$.
Then, we obtain that
\begin{align*}
nS'_j(X^n,Y^n) =&
\begin{cases}
-\sum_{i=1}^n \ln f(x_i|b^*) & j=1\\
-N_1\ln a -N_2\ln (1-a) - \sum_{i=1}^n \ln f(x_i|b^*) & j=2\\
-\sum_{i=1}^n \ln f(x_i|b^*) & j=3,
\end{cases}
\end{align*}
where $N_1=\sum_{i=1}^n \delta_{y_i,1}$ and $N_2=\sum_{i=1}^n \delta_{y_i,2}$.
The cases $j=1$ and $j=3$ have the same value
and the case $j=2$ is smaller than the others due to the first two terms in Eq.\ref{eq:S'_j},
which holds for any value of $0<a<1$ in $W^t_2$.
This means that $W^t_2$ cannot make $p(Y^n|X^n)$ maximum.
In other words, the assignment $Y^n$ using both labels $1$ and $2$ is not the peak.

\subsection{Two Approaches to Calculate $p(Y^n|X^n)$ and their Difference}
It is necessary to emphasize that the calculation of $p(Y^n|X^n)$
based on sampling from $p(w|X^n)$ following Eq. \ref{eq:learner_posterior} can be inaccurate.
According to Theorem \ref{th:PT_mixture} and Eq.\ref{eq:asymp_FX},
we confirm that $F(X^n)$ has a phase transition in mixture models
due to the hyperparameter of the Dirichlet prior.
This means that, when the hyperparameter $\eta_1$ is large,
the Monte Carlo sampling are from the area,
in which all the components are used such as $W^t_2$.
In the numerical computation,
the integrand of Eq. \ref{eq:learner_posterior} will be close to
$\prod_{i=1}^n p(x_i,y_i|w^*_2)/p(x_i|w^*_2)$, where $w^*_2\in W^t_2$.
Because $w^*_2\in W^t_2 \subset W^t_X$,
\begin{align*}
\prod_{i=1}^n \frac{p(x_i,y_i|w^*_2)}{p(x_i|w^*_2)}
=& \exp\bigg\{ \sum_{i=1}^n \ln p(x_i,y_i|w^*_2) - \sum_{i=1}^n \ln p(x_i|w^*_2)\bigg\}\\
=& \exp\big\{ -nS'_2(X^n,Y^n) + nS(X^n) \big\}.
\end{align*}
On the other hand, based on Eq. \ref{eq:learner},
the desired value of $p(Y^n|X^n)$ is calculated as
\begin{align*}
\frac{Z(X^n,Y^n)}{Z(X^n)} =& \exp\bigg\{ F(X^n)-F(X^n,Y^n)\bigg\}\\
=& \exp\{-nS(X^n,Y^n) + nS(X^n)\} +o(\exp(-n)).
\end{align*}
Since $S'_2(X^n,Y^n)>S(X^n,Y^n)$, the value of Eq. \ref{eq:learner_posterior}
is much smaller than that of Eq. \ref{eq:learner}.
Therefore, the result of the numerical integration in Eq. \ref{eq:learner_posterior} is almost zero.
The parameter area providing non-zero value of integrand in Eq. \ref{eq:learner_posterior}
is located in the tail of the posterior distribution when $p(w|X^n)$ converges to $W^t_X\setminus W^p_{XY}$.
\subsection{Failure of Parameter Sampling from the Posterior Distribution}
In the previous subsection, parameter sampling from the posterior distribution
can make an adverse effect on the calculation of the distribution of the latent variable.
Here, in the other way,
we show that latent-variable sampling can construct an undesired posterior distribution.

There are methods to sample a sequence of $\{w,Y^n\}$ from $p(w,Y^n|X^n)$.
Ignoring $Y^n$, we obtain the sequence $\{w\}$.
The Gibbs sampling in the MCMC method \cite{Robert2005} is one of the representative techniques.
\begin{quote}[Gibbs Sampling for a Model with a Latent Variable]
\begin{enumerate}
\item Initialize the parameter;
\item Sample $Y^n$ based on $p(Y^n|w,X^n)$;
\item Sample $w$ based on $p(w|Y^n,X^n)$;
\item Iterate by alternately updating Step 2 and Step 3.
\end{enumerate}
\end{quote}
The sequence of $\{w,Y^n\}$ obtained by this algorithm follows $p(w,Y^n|X^n)$.
The extracted parameter sequence $\{w\}$ is assumed to be samples from the posterior
because $p_G(w|X^n)=\sum_{Y^n} p(w,Y^n|X^n)$ is theoretically equal to $p(w|X^n)$.
However, in the mixture models,
the practical value of $p_G(w|X^n)$ based on the Monte Carlo method can be
different from that of the original posterior $p(w|X^n)$
when the hyperparameter for the mixing ratio $\eta_1$ is large.

Let us consider the expression
\begin{align*}
-\ln p(X^n,Y^n,w) =& -\ln \prod_{i=1}^n \prod_{k=1}^K a_k^{\delta_{y_ik}}f(x_i|b_k)^{\delta_{y_ik}} -\ln \varphi(w;\eta)\\
=& - \sum_{k=1}^K \delta_{y_ik} \ln a_k - \sum_{i=1}^n \sum_{k=1}^K \delta_{y_ik}\ln f(x_i|b_k) -\ln \varphi(w;\eta).
\end{align*}
We determine a location of a pair $(\bar{w},\bar{Y}^n)$ that minimizes this expression
in the asymptotic case $n\rightarrow \infty$
because the relation $p(X^n,Y^n,w)\propto p(w,Y^n|X^n)$
indicates that the sequence $\{w,Y^n\}$ is mainly taken from the neighborhood of the pair.
The third term of the last expression does not have any asymptotic effect
because it has the constant order on $n$.
The first two terms have the same expression as Eq. \ref{eq:S'_j}.
Based on the calculation of $S'_j(X^n,Y^n)$,
$\bar{w}\in W^p_{XY}$ and $\bar{Y}^n=\arg \max_{Y^n} p(X^n,Y^n,\bar{w})$.
Therefore, the practical value of $p_G(w|X^n)$ is calculated by the sequence $\{w\}$ around $W_{XY}^p$ for any $\eta_1$
while the convergence area of the original $p(w|X^n)$
depends on the phase of $F(X^n)$ controlled by $\eta_1$.

In Example \ref{ex:2comp_gaussian},
the posterior $p(w|X^n)$ converges to $W^t_2$ when $\eta_1$ is large.
On the other hand, the sampled sequence based on $p(X^n,Y^n,w)$ are mainly from $W_1\cup W_3$
since $S'_2(X^n,\bar{Y}^n_2)>S'_1(X^n,\bar{Y}^n_1)=S'_3(X^n,\bar{Y}^n_3)$,
where $\bar{Y}^n_j$ stands for the assignment minimizing $S'_j(X^n,Y^n)$.
In order to construct the sequence $\{w\}$ following $p(w|X^n)$,
we need samples $(w,Y^n)\in W_2\times \bar{Y}^n_2$,
which are located in the tail of $p(w,Y^n|X^n)$.
In theory, the sequence $\{w\}$ from $p(w,Y^n|X^n)$ realizes the one from $p(w|X^n)$.
However, in practice, it is not straightforward to obtain $\{w,Y^n\}$ from the tail of $p(w,Y^n|X^n)$.
This property of the Gibbs sampling has been reported in a Gaussian mixture model \cite{Nagata2008}.
The experimental results show that the obtained sequence of $\{w\}$ is localized in the area
corresponding to $W_{XY}^p$.
Note that there is no failure of the MCMC method when $\eta_1$ is sufficiently small,
where the peaks of $p(w|X^)$ and $p(w,Y^n|X^n)$ are in the same area.
Thus, to judge the reliability of the MCMC sampling,
we have to know the phase transition point such as $\eta_1=1/2$ in Lemma \ref{lem:binomial_D}.
\section{Conclusions}
\label{sec:Conc}
The present paper clarifies the asymptotic accuracy
of the Bayes latent-variable estimation.
The dominant order is at most $\ln n/n$, and its coefficient
is determined by a positional relation between the largest poles of the zeta functions.
According to the mixture-model case,
it is suggested that the order is dominant
and the coefficient is affected by the redundancy of the learning model and the hyperparameters.
The accuracy of prediction can be approximated by methods
such as the cross-validation and bootstrap methods.
On the other hand, there is no approximation for the accuracy of latent-variable estimation,
which indicates that the theoretical result plays a central role
in evaluating the model and the estimation method.
%
%
\section*{Appendix A.}
Here, we prove Theorems \ref{th:main} and \ref{th:error_mixture}, and Corollary \ref{cor:bound_gaussian}.
\subsection*{Proof of Theorem \ref{th:main}}
{\bf Proof:}
Let us define another average free energy as
\begin{align*}
F_{XY}(n) = -nS_{XY} +E_{XY}\bigg[-\ln Z(X^n,Y^n) \bigg],
\end{align*}
where the entropy function is given by
\begin{align*}
S_{XY} = - \sum_{y=1}^{K} \int q(x,y)\ln q(x,y)dx.
\end{align*}
According to the definitions of the error function $D(n)$
and the Bayes estimation method Eq. \ref{eq:learner}, it holds that
\begin{align*}
nD(n) =& E_{XY}\bigg[\ln\frac{q(X^n,Y^n)}{Z(X^n,Y^n)}\bigg]-E_X\bigg[\ln\frac{q(X^n)}{Z(X^n)}\bigg]\nonumber\\
=&-nS_{XY} -E_{XY}\bigg[\ln Z(X^n,Y^n)\bigg] +nS_X +E_X\bigg[\ln Z(X^n)\bigg]\nonumber\\
=&F_{XY}(n) - F_X(n).
\end{align*}
Based on (A1), (A2), and algebraic geometrical analysis,
we obtain the asymptotic forms of $F_{XY}(n)$ and $F_X(n)$:
\begin{align*}
F_{XY}(n) =& \lambda_{XY}\ln n - (m_{XY}-1)\ln\ln n +O(1),\\
F_X(n) =& \lambda_X \ln n -(m_X-1)\ln\ln n +O(1),
\end{align*}
which proves the theorem.
{\bf (End of Proof)}
\subsection*{Outline of the Calculation of a Pole of the Zeta Function}
\label{sec:outline}
We will show the outline of calculation to find a pole.
Let us introduce some useful lemmas for the zeta function.
The proofs are omitted because they are almost obvious due to the relation between
the free energy and the zeta function.
\begin{lemma}
\label{lem:inequality}
Let the largest poles of the zeta functions $\int H_1(w)^z\varphi(w)dw$
and $\int H_2(w)^z\varphi(w)dw$ be $z=-\lambda_1$ and $z=-\lambda_2$, respectively.
It holds that $\lambda_1\le \lambda_2$ when $H_1(w)\le H_2(w)$ on the support of $\varphi(w)$.
\end{lemma}
\begin{lemma}
\label{lem:equivalence}
Under the same conditions as Lemma \ref{lem:inequality},
it holds that $\lambda_1=\lambda_2$ if there exist positive constants $C_1$ and $C_2$
such that $C_1 H_2(w) \le H_1(w) \le C_2 H_2(w)$.
\end{lemma}
We define an equivalence relation $H_1(w)\equiv H_2(w)$
due to $\lambda_1=\lambda_2$ in Lemma \ref{lem:equivalence}.

Let us now calculate a general zeta function $\int H(w)^z\varphi(w)dw$.
First, we focus on the restricted area $W_{res}$,
which is the neighborhood of $\{w: H(w)=0\}$ in the parameter space
because poles of the zeta function do not depend on other areas \cite{Watanabe01a}.
Next, we need a function $H^{\text{alg}}(w)$, which is a polynomial of $w$
and satisfies $H(w)\equiv H^{\text{alg}}(w)$.
Based on Lemma \ref{lem:equivalence}, the largest pole of the zeta function
$\int_{W_{res}} H^{\text{alg}}(w)^z \varphi(w)dw$ is the same as that of the original zeta function.
According to the resolution of singularities \cite{Hironaka},
there is a mapping $u=\Phi(w)$ such that
\begin{align}
H^{\text{alg}}(\Phi(w)) =& a(u)u_1^{2\alpha_1}u_2^{2\alpha_2}\dots u_d^{2\alpha_d},\label{eq:resolution}
\end{align}
where $a(u)$ is a non-zero analytic function in $\{u:H(\Phi(w))=0\}$, and $\alpha_1,\dots,\alpha_d$ are integers.
Let $|\Phi|=|u_1|^{\beta_1}\dots |u_d|^{\beta_d}$ be the Jacobian,
and the prior distribution is described as $\varphi(\Phi(w))=u_1^{\gamma_1}\dots u_d^{\gamma_d}$,
where $\beta_i$ and $\gamma_i$ are integers.
Then, it holds that
\begin{align*}
\int_{W_{res}} H^{\text{alg}}(w)^z \varphi(w)dw
=& \int_{\Phi(W_{res})} H^{\text{alg}}(\Phi(w))^z \varphi(\Phi(w))|\Phi|du\\
=& \int_{\Phi(W_{res})} a(u)\prod_{i=1}^d u_i^{2\alpha_iz+\gamma_i}|u_i|^{\beta_i}du.
\end{align*}
Calculating the integral over $u_i$ in the last expression,
we find that the zeta function has factors $(2\alpha_iz+\beta_i+\gamma_i+1)^{-1}$.
This means that there are poles $z=-(\beta_i+\gamma_i+1)/(2\alpha_i)$.

When it is not straightforward to find the multiple form such as Eq.\ref{eq:resolution},
we can consider a partially-multiple form;
\begin{align*}
H^{\text{alg}}(\Phi(w)) = a(u) u_1^{2\alpha_1}g(u \setminus u_1),
\end{align*}
where the function $g(u\setminus u_1)$ can be a polynomial of $u\setminus u_1$.
The zeta function is written as
\begin{align*}
\int_{\Phi(w)} a(u)g(u\setminus u_1)u_1^{2\alpha_1z+\gamma_1}|u_1|^{\beta_1}du.
\end{align*}
Calculating the integral over $u_1$, we obtain a pole $z=-(\beta_1+\gamma_1+1)/(2\alpha_1)$.

Assume that we obtain a partially-multiple form as the upper bounds such that
\begin{align*}
H^{\text{alg}}(w)\le a(u)u_1^{2\alpha_1}g(u\setminus u_1),
\end{align*}
where the Jacobian and the prior include factors $|u_1|^{\beta_1}$ and $u_1^{\gamma_1}$, respectively.
Due to Lemma \ref{lem:inequality},
a pole of the zeta function with respect to the right-hand side provides the upper bounds
$\lambda \le (\beta_1+\gamma_1+1)/(2\alpha_1)$.
\subsection*{Proof of Theorem \ref{th:error_mixture}}
The following lemma shows the calculation of $\lambda_{XY}$.
\begin{lemma}
\label{lem:lambdaXY}
The largest pole of the zeta function $\zeta_{XY}(z)$ is
\begin{align*}
\lambda_{XY} =& \frac{K^*-1+K^* d_c}{2} + (K-K^*)\eta_1,\\
m_{XY} =& 1.
\end{align*}
\end{lemma}
{\bf Proof of Lemma \ref{lem:lambdaXY}:}
We consider a restricted parameter space $W_1$, which is a neighborhood of $W^t_{XY}$ given by
\begin{align*}
a_k =& a^*_k \;\; (2 \le k \le K^*),\\
a_k =& 0 \;\; (k> K^*),\\
b_k =& b^*_k \;\; (1 \le k \le K^*).
\end{align*}
This is a generalization of $W_1$ in Example \ref{ex:2comp_gaussian}.
The Kullback-Leibler divergence has the expression
\begin{align*}
H_{XY}(w) =& \sum_{k=1}^{K^*}a^*_k\bigg\{ \ln\frac{a^*_k}{a_k} + \int f(x|b^*_k)\ln \frac{f(x|b^*_k)}{f(x|b_k)}dx \bigg\} \\
\equiv& \bigg(1-\sum_{k=2}^{K^*}a^*_k\bigg)\ln\frac{1-\sum_{k=2}^{K^*}a^*_k}{1-\sum_{k=2}^Ka_k} + \int f(x|b^*_1)\ln\frac{f(x|b^*_1)}{f(x|b_1)}dx \\
&+ \sum_{k=2}^{K^*}\bigg\{ a^*_k\ln\frac{a^*_k}{a_k} + \int f(x|b^*_k)\ln\frac{f(x|b^*_k)}{f(x|b_k)}dx \bigg\}.
\end{align*}
Based on the shift transformation $\Phi_1(w)$, such that
\begin{align*}
\bar{a}_k =& a_k - a^*_k \;\; (2 \le k \le K^*),\\
\bar{a}_k =& a_k \;\; (k>K^*),\\
\bar{b}_{km} =& b_{km} - b^*_{km} \;\; (1 \le k \le K^*, 1\le m\le d_c),\\
\bar{b}_{km} =& b_{km} \;\; (k>K^*, 1\le m \le d_c),
\end{align*}
we can find an equivalent polynomial described as
\begin{align}
H_{XY}(\Phi_1(w)) \equiv& \sum_{k=2}^{K^*} \bar{a}_k^2 + \sum_{k=K^*+1}^K \bar{a}_k 
+ \sum_{k=1}^{K^*} \bar{b}_k^2.\label{eq:algHXY},
\end{align}
where the detailed derivation is in the Appendix B.
Let the right-hand side of Eq. \ref{eq:algHXY} be $H^{\text{alg}}_{XY}(\Phi_1(w))$, and
consider a zeta function given by
\begin{align*}
\zeta_1(z) = \int H^{\text{alg}}_{XY}(\Phi_1(w))^z\varphi(\Phi_1(w);\eta)d\Phi_1(w).
\end{align*}
According to Lemma \ref{lem:equivalence}, the positions of the poles of $\zeta_1(z)$ are the same as those of $\zeta_{XY}(z)$.
By using a blow-up $\Phi_2$ defined by
\begin{align*}
u_2 =& \bar{a}_2,\\
u_2u_k =& \bar{a}_k \;\; (2< k \le K^*),\\
u_2^2u_k =& \bar{a}_k \;\; (k>K^*),\\
u_2v_{km} =& \bar{b}_{km} \;\; (1\le k \le K^*, 1\le m \le d_c),\\
v_{km} =& \bar{b}_{km} \;\; (k>K^*, 1 \le m \le d_c),
\end{align*}
we obtain the following expression in the restricted area,
\begin{align*}
&\zeta_1(z) \\ &= \int_{\Phi_2\Phi_1(W_1)} f_1(\Phi_2\Phi_1(w))
u_2^{2z}\varphi(\Phi_2\Phi_1(w);\eta)|u_2|^{K^*-2+K^*d_c+2(K-K^*)}d\Phi_2\Phi_1(w),
\end{align*}
where $f_1$ is a function consisting of the parameters except for $u_2$,
and a factor on $|u_2|$ is derived from the Jacobian of $\Phi_2$.
Note that there is not $u_1$ as a parameter in $\Phi_2\Phi_1(w)$
since $w_1$ is already omitted on the basis of the relation $a_1=1-\sum_{k=2}^K a_k$.
The symmetric Dirichlet prior has a factor $\prod_{k=2}a_k^{\eta_1-1}$ in the original parameter space.
According to $\Phi_2\Phi_1(a_k)=u_2^2u_k$ for $k>K^*$,
it has a factor $u_2^{2(K-K^*)(\eta_1-1)}$ in the space of $\Phi_2\Phi_1(w)$,
which indicates that $\zeta_1(z)$ has a pole at $z=-(K^*-1+K^*d_c)/2-(K-K^*)\eta_1$.
Considering the symmetry of the parameters in $H_{XY}^{\text{alg}}(w)$,
we determine that this pole is the largest and that its order is $m_{XY}=1$,
which proves Lemma \ref{lem:lambdaXY}.
{\bf (End of Proof)}

The result for $\lambda_X$ is shown in the following lemma.
\begin{lemma}
\label{lem:lambdaX}
The largest pole of the zeta function $\zeta_X(z)$ has the bound
\begin{align*}
\lambda_X \le \mu =& \frac{K^*-1+K^* d_c}{2} + \frac{(K-K^*)\eta_1}{2}.\\
\end{align*}
\end{lemma}
{\bf Proof of Lemma \ref{lem:lambdaX}:}
It is known (cf. \cite{Yamazaki10a} ; Section 7.8 of \cite{Watanabe09:book})
that, in the restricted area $W_1$, there are positive constants $C_1$ and $C_1'$ such that
\begin{align*}
H_X(w) \le& C_1 \int \bigg\{\frac{p(x|w)}{q(x)}-1\bigg\}^2 dx\\
\equiv& C_1' \int\bigg\{ p(x|w)-q(x)\bigg\}^2 dx.
\end{align*}
Using $\Phi_1(w)$, we obtain 
\begin{align*}
H_X(\Phi_1(w)) \le & C_1' \int \bigg\{ \sum_{k=2}^{K^*} \bar{a}_k\big( f(x|\bar{b}_k+b^*_k)-f(x|\bar{b}_1+b^*_1)\big)\\
& + \sum_{k=2}^{K^*} a^*_k \big(f(x|\bar{b}_k+b^*_k)-f(x|b^*_k)\big)\\
& + \big(1-\sum_{k=2}^{K^*}a^*_k\big)\big(f(x|\bar{b}_1+b^*_1)-f(x|b^*_1)\big)\\
& + \sum_{k>K^*}^K\bar{a}_k\big(f(x|\bar{b}_k)-f(x|\bar{b}_1+b^*_1)\big)\bigg\}^2 dx.
\end{align*}
Because $f(x|b_k)$ is a regular model, there is a positive constant $C_2$ such that
\begin{align}
H_X(\Phi_1(w)) \le& C_2 \bigg\{ \sum_{k=2}^{K^*}\bar{a}_k^2 + \sum_{k=2}^{K^*}\bar{b}_k^2
+ \bar{b}_1^2 + \sum_{k>K^*}^K \bar{a}_k^2 \bigg\} \label{eq:algHX}
\end{align}
in $W_1$, where the detailed derivation is in the Appendix B.
Let the right-hand side be $H_X^{\text{alg}}(\Phi_1(w))$, and consider a zeta function given by
\begin{align*}
\zeta_2(z) = \int_{\Phi_1(W_1)} H_X^{\text{alg}}(\Phi_1(w))^z \varphi(\Phi_1(w);\eta)d\Phi_1(w).
\end{align*}
According to Lemma \ref{lem:inequality}, a pole $z=-\mu$ of the zeta function $\zeta_2(z)$
provides bounds for the largest pole of $\zeta_X(z)$, such that $z=-\lambda_X\ge-\mu$.
By using a blow-up $\Phi_3$ defined by
\begin{align*}
u_2 =& \bar{a}_2,\\
u_2u_k =& \bar{a}_k \;\; (2< k \le K^*),\\
u_2u_k =& \bar{a}_k \;\; (k>K^*),\\
u_2v_{km} =& \bar{b}_{km} \;\; (1\le k \le K^*, 1\le m \le d_c),\\
v_{km} =& \bar{b}_{km} \;\; (k>K^*, 1 \le m \le d_c),
\end{align*}
we obtain 
\begin{align*}
&\zeta_2(z)\\ &= \int_{\Phi_3\Phi_1(W_1)} f_2(\Phi_3\Phi_1(w))
u_2^{2z}\varphi(\Phi_3\Phi_1(w);\eta)|u_2|^{K^*-2+K^*d_c+(K-K^*)}d\Phi_3\Phi_1(w),
\end{align*}
where $f_2$ is a function of the parameters except for $u_2$,
and the factor on $|u_2|$ is derived from the Jacobian of $\Phi_3$.
It is easy to confirm that the Dirichlet prior has a factor $u_2^{(K-K^*)(\eta_1-1)}$.
Therefore, $\zeta_2(z)$ has a pole at $z=-\mu=-(K^*-1+K^*d_c)/2-(K-K^*)\eta_1/2$,
which proves Lemma \ref{lem:lambdaX}.
{\bf (End of Proof)}

We are now prepared to prove Theorem \ref{th:error_mixture}.
{\bf Proof of Theorem \ref{th:error_mixture}:}
According to Theorem \ref{th:main}, it holds that
\begin{align*}
D(n) =& (\lambda_{XY}-\lambda_X)\frac{\ln n}{n} + o\bigg(\frac{\ln n}{n}\bigg).
\end{align*}
Combining Lemmas \ref{lem:lambdaXY} and \ref{lem:lambdaX}, we obtain 
\begin{align*}
D(n) \ge& \bigg\{\frac{K^*-1+K^*d_c}{2}+(K-K^*)\eta_1\\
& -\frac{K^*-1+K^*d_c}{2}-\frac{(K-K^*)\eta_1}{2}\bigg\}\frac{\ln n}{n} +o\bigg(\frac{\ln n}{n}\bigg)\\
= & \frac{(K-K^*)\eta_1}{2}\frac{\ln n}{n} + o\bigg(\frac{\ln n}{n}\bigg),
\end{align*}
which completes the proof.
{\bf (End of Proof)}
\subsection*{Proof of Corollary \ref{cor:bound_gaussian}}
{\bf Proof:}
Since $H_X(w)$ has the bound,
\begin{align}
H_X(w) \le& C_1\int \bigg(\frac{p(x|w)}{q(x)}-1\bigg)^2 dx, \label{eq:quad_bound}
\end{align}
Lemma \ref{lem:lambdaX} immediately holds.
Due to the analytic divergence $H_{XY}(w)$, Lemma \ref{lem:lambdaXY} also holds.
Combining these lemmas, we obtain the same lower bound as Theorem \ref{th:error_mixture}.
In the Gaussian mixture,
\begin{align*}
H_{XY}(w) =& \sum_{y=1}^{K^*} \int q(x,y) \ln\frac{a^*_y f(x|b^*_y)}{a_y f(x|b_y)}dx\\
=& \sum_{y=1}^{K^*} a^*_y \ln\frac{a^*_y}{a_y} + \sum_{y=1}^{K^*} a^*_y f(x|b^*_y) \ln \frac{f(x|b^*_y)}{f(x|b_y)}dx.
\end{align*}
Because $f(x|b)$ is identifiable, $H_{XY}(w)$ is analytic.
Section 7.8 in \cite{Watanabe09:book}
shows that $H_X(w)$ has the upper bound expressed as Eq.\ref{eq:quad_bound} in the Gaussian mixture,
which proves the corollary.
{\bf (End of Proof)}
\section*{Appendix B.}
This section shows supplementary proofs for some equations
in the proof of Theorem \ref{th:error_mixture}.

According to the analysis with the Newton diagram \cite{Yamazaki10a},
the following relations hold;
\begin{align}
w_1 + \big\{ h_0(w\setminus w_1) + w_1h_1(w)\big\}^2 \equiv& w_1 + h_0(w\setminus w_1)^2, \label{eq:rel1}\\
w_1^2 + \big\{ h_0(w\setminus w_1) + w_1h_1(w)\big\}^2 \equiv & w_1^2 + h_0(w\setminus w_1)^2, \label{eq:rel2}\\
w_1 + w_1h_1(w) \equiv& w_1, \label{eq:rel3}\\
w_1^2 + w_1^2h_1(w) \equiv& w_1^2, \label{eq:rel4}
\end{align}
where $w=\{w_1,w_2,\dots,w_d\}$, and $h_0$ and $h_1$ are polynomial.
Using these relations, we prove Eqs.\ref{eq:algHXY} and \ref{eq:algHX}.
\subsection*{Proof of Equation \ref{eq:algHXY}}
Recall that the Kullback-Leibler divergence has the following equivalent expression;
\begin{align*}
H_{XY}(w) =& \sum_{k=1}^{K^*}a^*_k\bigg\{ \ln\frac{a^*_k}{a_k} + \int f(x|b^*_k)\ln \frac{f(x|b^*_k)}{f(x|b_k)}dx \bigg\} \\
\equiv& \bigg(1-\sum_{k=2}^{K^*}a^*_k\bigg)\ln\frac{1-\sum_{k=2}^{K^*}a^*_k}{1-\sum_{k=2}^Ka_k} + \int f(x|b^*_1)\ln\frac{f(x|b^*_1)}{f(x|b_1)}dx \\
&+ \sum_{k=2}^{K^*}\bigg\{ a^*_k\ln\frac{a^*_k}{a_k} + \int f(x|b^*_k)\ln\frac{f(x|b^*_k)}{f(x|b_k)}dx \bigg\}.
\end{align*}

Based on the transformation $\Phi_1(w)$
and the Taylor expansion of $\ln(1+\Delta x)$ around $|\Delta x|= 0$, we obtain
\begin{align*}
H_{XY}(\Phi_1(w)) \equiv& - \bigg(1-\sum_{k=2}^{K^*}a^*_k\bigg)\ln\bigg(1-\sum_{k=2}^K\frac{\bar{a}_k}{1-\sum_{k=2}^{K^*}a^*_k}\bigg)\\
&+ \int f(x|b^*_1)\ln\frac{f(x|b^*_1)}{f(x|b^*_1+\bar{b}_1)}dx\\
&+ \sum_{k=2}^{K^*}\bigg\{ -a^*_k\ln\bigg(1+\frac{\bar{a}_k}{a^*_k}\bigg) + \int f(x|b^*_k)\ln\frac{f(x|b^*_k)}{f(x|b^*_k+\bar{b}_k)}dx \bigg\}\\
\equiv& \sum_{k=2}^K \bar{a}_k - \sum_{k=2}^{K^*} \bar{a}_k\\
& -\frac{1}{2}\bigg(1-\sum_{k=2}^{K^*}a^*_k\bigg)^{-1}\bigg(\sum_{k=2}^K \bar{a}_k\bigg)^2
+\frac{1}{2}\sum_{k=1}^{K^*}a_k^{*-1}\bar{a}_k^2 + h_r(w)\\
&+ \sum_{k=1}^{K^*} \int f(x|b^*_k)\ln\frac{f(x|b^*_k)}{f(x|b^*_k+\bar{b}_k)}dx,
\end{align*}
where $h_r(w)$ includes the higher order terms on $\bar{a}_k$.
By applying Eq.\ref{eq:rel2} to $\bar{a}_k^2$, it holds that
\begin{align*}
H_{XY}(\Phi_1(w)) \equiv& \sum_{k=2}^K \bar{a}_k - \sum_{k=2}^{K^*} \bar{a}_k\\
& +\frac{1}{2}\sum_{k=1}^{K^*}a_k^{*-1}\bar{a}_k^2 + h_r(w)\\
&+ \sum_{k=1}^{K^*} \int f(x|b^*_k)\ln\frac{f(x|b^*_k)}{f(x|b^*_k+\bar{b}_k)}dx\\
=& \sum_{k=K^*+1}^K \bar{a}_k +\frac{1}{2}\sum_{k=1}^{K^*}a_k^{*-1}\bar{a}_k^2 + h_r(w)\\
&+ \sum_{k=1}^{K^*} \int f(x|b^*_k)\ln\frac{f(x|b^*_k)}{f(x|b^*_k+\bar{b}_k)}dx.
\end{align*}
Due to Eqs. \ref{eq:rel3} and \ref{eq:rel4}, $h_r(w)$ is excluded;
\begin{align*}
H_{XY}(\Phi_1(w)) \equiv& \sum_{k=2}^{K^*} \bar{a}_k^2 + \sum_{k=K^*+1}^K \bar{a}_k 
+ \sum_{k=1}^{K^*} \int f(x|b^*_k)\ln\frac{f(x|b^*_k)}{f(x|b^*_k+\bar{b}_k)}dx.
\end{align*}
Because $f(x|b_k)$ is regular, it is known that
\begin{align*}
\int f(x|b^*_k)\ln\frac{f(x|b^*_k)}{f(x|b^*_k+\bar{b}_k)}dx \equiv \bar{b}_k^2,
\end{align*}
which proves that
\begin{align*}
H_{XY}(\Phi_1(w)) \equiv& \sum_{k=2}^{K^*} \bar{a}_k^2 + \sum_{k=K^*+1}^K \bar{a}_k 
+ \sum_{k=1}^{K^*} \bar{b}_k^2.
\end{align*}
{\bf (End of Proof)}
\subsection*{Proof of Equation \ref{eq:algHX}}
Recall that the Kullback-Leibler divergence $H_X(\Phi_1(w))$ has the following bound;
\begin{align*}
H_X(\Phi_1(w)) \le & C_1' \int \bigg\{ \sum_{k=2}^{K^*} \bar{a}_k\big( f(x|\bar{b}_k+b^*_k)-f(x|\bar{b}_1+b^*_1)\big)\\
& + \sum_{k=2}^{K^*} a^*_k \big(f(x|\bar{b}_k+b^*_k)-f(x|b^*_k)\big)\\
& + \big(1-\sum_{k=2}^{K^*}a^*_k\big)\big(f(x|\bar{b}_1+b^*_1)-f(x|b^*_1)\big)\\
& + \sum_{k>K^*}^K\bar{a}_k\big(f(x|\bar{b}_k)-f(x|\bar{b}_1+b^*_1)\big)\bigg\}^2 dx.
\end{align*}
In the area $\Phi_1(W_1)$, there is a positive constant $C''_1$ such that
\begin{align}
H_X(\Phi_1(w)) \le & C_1'' \bigg\{
\sum_{k=1}^{K^*} \bar{a}_k^2 \int\big( f(x|\bar{b}_k+b^*_k)-f(x|\bar{b}_1+b^*_1)\big)^2dx\nonumber\\
& + \sum_{k=1}^{K^*} \int\big(f(x|\bar{b}_k+b^*_k)-f(x|b^*_k)\big)^2dx + \sum_{k>K^*}^K\bar{a}_k^2 \bigg\}. \label{eq:C1dd}
\end{align}
The Taylor expansion at $\bar{b}_k$ yields
\begin{align*}
f(x|\bar{b}_k+b^*_k) =& f(x|b^*_k) + \bar{b}_k^\top\frac{\partial}{\partial \bar{b}_k}f(x|b^*_k) +\dots.
\end{align*}
The second term of the right-hand side in Eq.\ref{eq:C1dd} has the following bound,
\begin{align*}
\sum_{k=1}^{K^*} \int\big(f(x|\bar{b}_k+b^*_k)-f(x|b^*_k)\big)^2dx \le C_b \sum_{k=1}^{K^*}
\bigg\{ \bar{b}_k^2 + \bar{b}_k^2 h_r(\bar{b}_k)\bigg\},
\end{align*}
where $C_b$ is a positive constant and $\bar{b}_k^2h_r(\bar{b}_k)$ stands for the rest of the terms.
Based on Eq.\ref{eq:rel4}, the bound has the equivalent form,
\begin{align*}
\sum_{k=1}^{K^*} \bigg\{ \bar{b}_k^2 + \bar{b}_k^2 h_r(\bar{b}_k)\bigg\} \equiv \sum_{k=1}^{K^*} \bar{b}_k^2,
\end{align*}
which changes the first term of Eq.\ref{eq:C1dd} into
\begin{align*}
\bar{a}_k^2 \int\big( f(x|\bar{b}_k+b^*_k)-f(x|\bar{b}_1+b^*_1)\big)^2dx \equiv \bar{a}_k^2
\end{align*}
due to Eq.\ref{eq:rel2}.
Then, there is a positive constant $C_2$ such that
\begin{align*}
H_X(\Phi_1(w)) \le& C_2 \bigg\{ \sum_{k=2}^{K^*}\bar{a}_k^2 + \sum_{k=2}^{K^*}\bar{b}_k^2
+ \bar{b}_1^2 + \sum_{k>K^*}^K \bar{a}_k^2 \bigg\}.
\end{align*}
{\bf (End of Proof)}
\section*{Appendix C.}
\subsection*{Proof of Lemma \ref{lem:binomial_D}}
{\bf Proof:}
The calculation is based on the way of the proof of Theorem \ref{th:error_mixture}.
Define the shift transformation $\Phi_4$ given by
\begin{align*}
\bar{a} =& 1 - a, \\
\bar{b}_1 =& (1-\bar{a})(b_1 - b^*) + \bar{a}\bar{b}_2, \\
\bar{b}_2 =& b_2 - b^*.
\end{align*}
This corresponds to focusing on the area $W_1\cup W_2$.
Following the calculation of \cite{Yamazaki10a},
we obtain
\begin{align*}
H_X(\Phi_4(w))\equiv& \bar{b}_1^2 + \bar{a}^2\bar{b}_2^4.
\end{align*}
Let the right-hand side be $H_{X2}^{\text{alg}}(w)$, and consider a zeta function given by
\begin{align*}
\zeta_3(z) =& \int_{W_1\cup W_2} H_{X2}^{\text{alg}}(\Phi_4(w))^z \varphi(\Phi_4(w);\eta)d\Phi_4(w).
\end{align*}
By using a blow-up $\Phi_5$ defined by
\begin{align*}
\bar{a} =& v_1v_2,\\
\bar{b}_1 =& u_1^2v_1,\\
\bar{b}_2 =& u_1,
\end{align*}
we obtain the following expression,
\begin{align*}
\zeta_3(z) =& \int_{\Phi_5\Phi_4(W_1\cup W_2)} f_3(\Phi_5\Phi_4(w))
u_1^{4z}v_1^{2z}\varphi(\Phi_5\Phi_4(w);\eta)|u_1|^2|v_1|d\Phi_5\Phi_4(w),
\end{align*}
where $f_3$ is a function of the parameter $v_2$.
The prior has a factor $v_1^{\eta_1-1}$.
Therefore, $\zeta_3(z)$ has poles at $z=-3/4$ and $z=-(1+\eta_1)/2$,
which are calculated from the factors $u_1$ and $v_1$, respectively.
Considering the cases $u_1=0$ and $v_1=0$,
we find that the effective area of the pole $z=-3/4$ is $W_2$
and that of $z=-(1+\eta_1)/2$ is $W_1$.
Due to the symmetry, the area $W_2\cup W_3$ has the same poles.
Then, the largest pole changes at $\eta_1=1/2$,
where the order of the pole is $m_X=2$.
This completes the proof.
{\bf (End of Proof)}
\subsection*{Proof of Theorem \ref{th:PT_mixture}}
First, we introduce tighter upper bounds on $\lambda_X$.
\begin{lemma}
\label{lem:tight_bounds}
Under the same condition as in Theorem \ref{th:error_mixture},
it holds that
\begin{align*}
\lambda_X \le& 
\begin{cases}
\frac{K^*-1+K^* d_c}{2} + \frac{K-K^*}{2}\eta_1 & \eta_1 \le d_c,\\
\frac{K^*-1+K^* d_c}{2} + \frac{(K-K^*)d_c}{2} & \eta_1 > d_c.
\end{cases}
\end{align*}
\end{lemma}
{\bf Proof:}
Consider the area $W_2$, which is the neighborhood of
\begin{align*}
a_k =& a^*_k \;\; (2\le k\le K^*)\\
b_{km} =& b^*_{km} \;\; (1\le k\le K^*, 1\le m \le d_c)\\
b_{km} =& b^*_{1m} \;\; (k>K^*, 1\le m \le d_c).
\end{align*}
Let us define the shift transformation $\Phi_5$ given by
\begin{align*}
\bar{a}_k =& a_k - a^*_k \;\; (2\le k\le K^*)\\
\bar{b}_{km} =& b_{km} - b^*_{km} \;\; (1\le k\le K^*, 1\le m \le d_c)\\
\bar{b}_{km} =& b_{km} - b^*_{1m} \;\; (k>K^*, 1\le m \le d_c).
\end{align*}
Based on the Taylor expansion of $f(x|\bar{b}_k+b^*_k)$, there is a positive constant $C_3$ such that
\begin{align*}
H_X(\Phi_5(w))\le C_3 \bigg\{ \sum_{k=2}^{K^*} \bar{a}_k^2 + \sum_{k=1}^K \bar{b}_k^2\bigg\}.
\end{align*}
Let the right-hand side be $H_{X3}^{\text{alg}}(w)$, and consider a zeta function given by
\begin{align*}
\zeta_4(z) =& \int_{\Phi_6(W_{21})} H_{X3}^{\text{alg}}(\Phi_6(w))^z \varphi(\Phi_6(w);\eta)d\Phi_6(w).
\end{align*}
By using a blow-up $\Phi_7$ defined by
\begin{align*}
u_2 =& \bar{a}_2,\\
u_2u_k =& \bar{a}_k \;\; (2\le k\le K^*),\\
u_k =& \bar{a}_k \;\; (k>K^*),\\
u_2v_{km} =& \bar{b}_{km} \;\; (1\le k\le K, 1\le m\le d_c),
\end{align*}
we obtain the following expression:
\begin{align*}
\zeta_4(z) =& \int_{\Phi_7\Phi_6(W_{21})} f_4(\Phi_7\Phi_6(w))
u_2^{2z}\varphi(\Phi_7\Phi_6(w);\eta)|u_2|^{K^*-2+Kd_c}d\Phi_7\Phi_6(w),
\end{align*}
where $f_4$ is a function consisting of the parameters except for $u_2$.
Therefore, $\zeta_4(z)$ has a pole at $z=-(K^*-1+Kd_c)/2$,
which shows that
\begin{align*}
\lambda_X \le& \frac{K^*-1+K^*d_c}{2} + \frac{(K-K^*)d_c}{2}.
\end{align*}
Compared to the result of Lemma \ref{lem:lambdaX},
we find that the bounds are tighter when $\eta_1>d_c$, which proves the lemma.
{\bf (End of Proof)}
Second, the following lemma shows the lower bound of $\lambda_X$;
\begin{lemma}
Under the same condition as in Theorem \ref{th:error_mixture},
it holds that
\begin{align*}
\label{lem:lower_bounds}
\lambda_X > \frac{K^*-1+K^* d_c}{2}.
\end{align*}
\end{lemma}
{\bf Proof:}
We can immediately obtain the inequality
based on the minimality condition of $q(x)$ and $d > K^*-1+K^* d_c$.
{\bf (End of Proof)}
Last, using these lemmas, we prove Theorem \ref{th:PT_mixture}.
As shown in the proofs of Lemmas \ref{lem:lambdaX} and \ref{lem:tight_bounds},
$\lambda_X$ is a linear function of $\eta_1$ due to the factor $a_k^{\eta_1-1}$ in the Dirichlet prior.
The upper and lower bounds imply that,
for $\eta_1$ close to zero, there exists a constant $\alpha$ such that
\begin{align*}
\lambda_X = \alpha \eta_1 + \beta,
\end{align*}
where $\beta=(K^*-1+K^* d_c)/2$.
Eliminated components appear in $\alpha \eta_1$
since their mixing ratio parameters converge to zero in the effective area,
and the prior factor $a_k^{\eta_1-1}$ works on the calculation of the pole of $\zeta_X(z)$.
The phase in the upper bounds eliminates all redundant components,
and the constant term $\beta$ in the above expression is the same value as that of the bounds.
This means that the redundant components are all eliminated in this phase.
On the other hand, the upper bounds also indicate that $\lambda_X$ must be a constant function
for a sufficiently large $\eta_1$.
When there is no linear factor of $\eta_1$ in $\lambda_X$,
all mixing ratio parameters converge to nonzero values;
all components are used in this phase.
Therefore, we have found the two phases, as desired.
{\bf (End of Proof)}

\bibliography{LearningTheory}
\bibliographystyle{plain}
\end{document}